\documentclass{article}
\usepackage{graphicx} 
\usepackage{xcolor}

\usepackage{amssymb,amsmath,amsthm,bbm}
\usepackage{verbatim,float,url,dsfont}
\usepackage{graphicx,subcaption,psfrag}
\usepackage{algorithm,algorithmic}
\usepackage{mathtools,enumitem}
\usepackage{multirow}
\usepackage{ragged2e}
\usepackage{xr-hyper}
\usepackage{array}

\usepackage[colorlinks=true,citecolor=blue,urlcolor=blue,linkcolor=blue]{hyperref}
\usepackage[margin=1in]{geometry}
\usepackage[round]{natbib}

\usepackage[utf8]{inputenc} 
\usepackage[T1]{fontenc}    
\usepackage{booktabs}       
\usepackage{nicefrac}         
\usepackage{microtype}      

\ifdefined\TimesFont 
\usepackage{times} 
\fi

\ifdefined\ParSkip 
\usepackage{parskip} 
\fi

\theoremstyle{definition}

\newtheorem*{assumption*}{\assumptionnumber}
\providecommand{\assumptionnumber}{}


\makeatletter
\newcommand*\rel@kern[1]{\kern#1\dimexpr\macc@kerna}
\newcommand*\widebar[1]{%
  \begingroup
  \def\mathaccent##1##2{%
    \rel@kern{0.8}%
    \overline{\rel@kern{-0.8}\macc@nucleus\rel@kern{0.2}}%
    \rel@kern{-0.2}%
  }%
  \macc@depth\@ne
  \let\math@bgroup\@empty \let\math@egroup\macc@set@skewchar
  \mathsurround\z@ \frozen@everymath{\mathgroup\macc@group\relax}%
  \macc@set@skewchar\relax
  \let\mathaccentV\macc@nested@a
  \macc@nested@a\relax111{#1}%
  \endgroup
}
\makeatother

\DeclareMathOperator*{\argmin}{argmin}

\def\P{\mathbb{P}}
\def\R{\mathbb{R}}

\def\cA{\mathcal{A}}

\def\cE{\mathcal{E}}

\def\cG{\mathcal{G}}

\def\cL{\mathcal{L}}

\def\cR{\mathcal{R}}

\def\cX{\mathcal{X}}
\def\cY{\mathcal{Y}}

\title{Unifying Image Counterfactuals and Feature Attributions with Latent-Space Adversarial Attacks}

\date{}
\author{Jeremy Goldwasser\thanks{Department of Statistics, University of California, Berkeley}
  \and 
  Giles Hooker\thanks{Department of Statistics and Data Science, University of Pennsylvania}}

\begin{document}

\maketitle
\begin{abstract}
    Counterfactuals are a popular framework for interpreting machine learning predictions. 
    These \textit{what if} explanations are notoriously challenging to create for computer vision models: standard gradient-based methods are prone to produce adversarial examples, in which imperceptible modifications to image pixels provoke large changes in predictions.
    We introduce a new, easy-to-implement framework for counterfactual images that can flexibly adapt to contemporary advances in generative modeling. Our method, \textit{Counterfactual Attacks}, resembles an adversarial attack on the representation of the image along a low-dimensional manifold. 
    In addition, given an auxiliary dataset of image descriptors, we show how to accompany counterfactuals with feature attribution that quantify the changes between the original and counterfactual images. 
    These importance scores can be aggregated into global counterfactual explanations that highlight the overall features driving model predictions.
    While this unification is possible for any counterfactual method, it has particular computational efficiency for ours. 
    We demonstrate the efficacy of our approach with the MNIST and CelebA datasets.
\end{abstract}
\section{Introduction}

In broad terms, a counterfactual statement is a \textit{what if}. It elucidates causal relationships by highlighting that if some factor had been different, then the outcome would have changed. This intuitive form of explanation was extended to machine learning (ML) by \citet{wachter2018counterfactual}. Like SHAP and LIME, counterfactuals interpret a model's prediction on an individual sample \citep{SHAP,LIME}. Their local explanations modify the input in a concise, realistic, actionable way that provokes a large change in the model output \citep{survey}. Ideally, these alterations illuminate the salient reason(s) behind the prediction. Such interpretations may empower humans to learn from ML models, build trust in them, or identify their failure modes.

Most prior works produce counterfactuals by solving a regularized regression problem. The loss encourages the model's output on the counterfactual to be near a pre-specified target, i.e. flipping the prediction. The regularization term penalizes examples that are too far from the original input. Convential notions of proximity include low overall deviation \citep{wachter2018counterfactual},  sparse interventions \citep{dandl2020multiobjective}, and plausible achievability \citep{asemota}. 

While tuning the regularization hyperparameter may be challenging or arbitrary, in general this is an effective approach for explaining tabular models. In fact, the counterfactual changes may even be interpreted as attribution scores \citep{cf_attributions}. Features whose values changed a lot for the counterfactual are deemed more important.

On image data, however, regularized optimization faces unique challenges. Neural networks are prone to adversarial attacks on computer vision tasks \citep{adversarial_attacks}. Using gradient ascent methods like FGSM, potentially imperceptible changes can yield wildly different outputs. Therefore it may be possible to construct counterfactuals with no substantive alteration. 

Making any sort of substantive edit to an image is challenging in its own right. Images are very high dimensional: 256x256 RGB images, for example, have roughly 200,000 pixels. Even without the presence of adversarial examples, blindly editing pixels will likely not correspond to any semantically meaningful change. Nor can these changes be induced by clever regularization. Sparsity penalities, for example, would render it impossible to learn counterfactuals with simple transformations in global features like lighting.

To address this, we propose new methodology to produce highly realistic image counterfactuals. Our approach, Counterfactual Attacks, runs gradient ascent on a latent representation of the image, in essence generating an adversarial example on the data manifold. 
Unlike previous methods, ours enables use of modern generative models like StyleGAN3 entirely off-the-shelf, and has no regularization hyperparameters to tune. 
Each counterfactual is the endpoint of a smooth trajectory starting with the original image; the intermediate images along this path may be visualized as well. 

Furthermore, our method can be unified with feature attributions when auxiliary labels are present. That is, the content of the counterfactual is captured in a set of importance scores which explain what changed from the original image. This unification allows counterfactuals to be analyzed in an automated fashion, for example to attain global importance scores without manually annotating hundreds of images. 
This concise summary provides novel insights into the salient factors driving model behavior. 
We show how to produce important scores not only for our algorithm, but for any counterfactual methodology.
To our knowledge, this is the first work quantifying image counterfactuals with feature importance scores.\footnote{Code for this work is available at \url{https://github.com/jeremy-goldwasser/counterfactual-attacks}.}


\section{Background}\label{sec:background}

Formally, let $x$ be a sample input to model $f$. The goal is to generate a similar, feasible input $x'$ whose output $f(x')$ is near the desired target $y'$. 

\subsection{Standard Counterfactual Approaches}

By definition, counterfactuals must balance multiple objectives. A loss objective such as squared error ensures $f(x')$ is close to $y'$. In addition, one or more proximity objectives ensures $x'$ is close to $x$.  For example, \citet{wachter2018counterfactual} proposed using a weighted L1 norm to measure the size of the change $x'-x$ that produces a desired change $f(x')$. Subsequently, \citet{dandl2020multiobjective} added a $L_0$ penalty for sparsity, as well as the distance to the nearest input for feasibility.

Some works simultaneously optimize the objectives via genetic algorithms, e.g. \citet{dandl2020multiobjective, dice, asemota}. More common, however, is the original framework for ML counterfactuals, introduced by \citet{wachter2018counterfactual}. In this setting, loss and regularizer objectives, $\cL$ and $\cR$, are combined into a hyperparameter-weighted sum. The expression is solved with gradient-based optimization to produce a counterfactual:

\begin{equation}\label{eq:reg}
    x' = \argmin_{\tilde x} \cL(f(\tilde{x}), y') + \lambda \cR(\tilde{x}, x).
\end{equation}

Another seminal paper, DiCE, proposes generating a set of counterfactuals, rather than just one \citep{dice}. The rationale for doing so is the so-called Rashomon Effect -- a given outcome may have multiple explanations, each valid in its own right but at odds with one another. In this case, different modifications to an input may cause the prediction to change in a desired direction.

\subsection{Image Counterfactuals}

A number of works have extended the notion of counterfactuals to images, leveraging powerful generative models. Several use Generative Adversarial Networks, or GANs \citep{goodfellow2014generative}. \citet{cond_gan_cf} trained a Conditional GAN that encodes an image, along with a desired amount to change its prediction on some model. The generator reconstructs the image, training to both fool the discriminator and shift the prediction accordingly.  
In contrast, the approach by \citet{gan_cf_like_us} ignores the predictor in question, training an unconditional DCGAN off-the-shelf \citep{DCGAN}. To generate counterfactuals, it solves an optimization problem akin to Equation \eqref{eq:reg} in the latent space. They use cross-entropy loss for classification error and a pixel-wise L1-norm for the penalty. 


In recent years, diffusion models have been shown to outperform GANs for image generation \citep{dhariwal2021diffusion}. Analogous to GANs, counterfactual works propose training conditional diffusion models and targeted modifications from unconditional models. For the former, \citet{diffusion_cond} trains a class-conditional diffusion model, sampling through the reverse process with altered classes to generate counterfactuals. They introduce various regularization techniques to control the stability of generation.
For the case of diffusion models, \citet{diffusion_class_guidance} simply train a unconditional Denoising Diffusion Probabilistic Model (DDPM) on the data, then sample a counterfactual using guided diffusion \citep{dhariwal2021diffusion}. 
In sampling through the reverse path, guided diffusion moves in the direction of the negative gradient of a loss function. Here, the loss is the hyperparameter-weighted sum of classification and proximity terms, again akin to Equation \eqref{eq:reg}.

While these methods are capable of generating high-quality counterfactuals, they suffer a number of drawbacks. Firstly, none of them guarantee the counterfactual will have the desired score. Rather, they either minimize regularized losses or direct sampling towards $y'$. In addition, they all have implementation challenges. GANs, for example, are notoriously difficult to train \citep{mescheder2018training}, and the methodology of \citet{diffusion_cond} is quite complicated. \citet{gan_cf_like_us} and \citet{diffusion_class_guidance} circumvent these issues somewhat by leveraging off-the-shelf models; however, both require tuning the regulization hyperparameter, which lacks reasonable heuristics in this context. Moreover, defining the perceptive loss is unintuitive. Pixel-wise loss, suggested by \citet{gan_cf_like_us}, restricts the space of possible counterfactuals, as in the aforementioned lighting example. 

A separate line of work, albeit further from counterfactuals, takes a different approach. It discovers latent image features in generative models like age and gender, then uses them to edit to an image. InterFaceGAN \citep{interpreting_linear} extracted the latent space representations of face images used to train PGGAN \citep{PGGAN} and StyleGAN \citep{stylegan}. With these relatively low-dimensional vectors, they fit linear models to features like age, glasses, and gender. These directions can then be used to edit images according to these features. 

The InterFaceGAN method relies on having labels for semantically meaningful image concepts.
While datasets like CelebA have such features, it is not true in general. To discover latent features, StylEx \citep{yossi_stylex} merely perturbs latent features one at a time in the StyleSpace of StyleGAN2 \citep{StyleGAN2}. This space appears after a few layers of transformations to the initial latent space in StyleGAN models. It has been shown to be highly disentangled, making StyleGAN a natural choice for image manipulation \citep{stylespace_rules}. 

\subsection{Dimensionality Reduction}

Our methodology rests upon the observation that neural networks are capable of representing data in a low-dimensional manifold. 
A number of architectures are capable of providing such encodings.
For example, data compression with autoencoders dates back decades \citep{lecun1987modeles,ballard1987modular,bourlard1988autoassociation, hinton1993autoencoders, hinton2006reducing}. VAEs \citep{VAEs} greatly improved upon vanilla autoencoders by encoding data into a smooth, probabilistic latent space. This enables the decoder to be more effectively used as a generative model. While autoencoders generally struggle to generate natural images, recent years have seen a number of significant advancements, e.g. Adversarial Autoencoders \citep{makhzani2016adversarialautoencoders}, VQ-VAE \citep{Oord2017NeuralDR}, and NVAE \citep{vahdat2021nvae}.

GANs also perform dimensionality reduction, and are more tailored for images. The latent code input to the generator has far lower dimension than the input. Similarly, the default implementation of the StyleGAN models produces a StyleSpace with dimension 512. 

The latent space of diffusion models like DDPMs tends to have the same dimensionality as the images, due to their foward and reverse noising process \citep{DDPM}. Therefore diffusion generally cannot be conceived as a dimensionality reduction technique. Latent Diffusion Models \citep{LDMs} and Diffusion Autoencoders \citep{preechakul2022diffusionautoencoders} are notable exceptions, and may be cast into our framework. 


Having trained an autoencoder, the low-dimensional representation for an image is obtained simply by running it through the encoder. Similarly, for LDMs and DAEs, the encoded image is then passed through the forward noising process. The task of ``GAN inversion,'' however, may be more challenging \citep{xia2023ganinversion}. Some GAN architectures have an encoder that maps to the generator, in which case the embedding may be obtained directly. Such an encoder may be trained jointly with the GAN itself, a technique first proposed independently by \citet{donahue2017adversarialfeaturelearning} and \citet{dumoulin2017adversariallylearnedinference}. Encoders may also be trained retrospectively, freezing the pre-trained weights of the generator and discriminator \citep{zhu2016generative, richardson2021encodingstylestyleganencoder}. Alternatively, encoder-free methods directly optimize the latent code for each image. For example, \citet{abdal2019image2stylegan} learns a StyleSpace vector whose reconstruction minimizes a loss with pixel-wise and perceptual terms; the latter term seeks to reproduce the internal activations of a pre-trained VGG model \citep{feifei}.


\section{Counterfactual Attacks}

Our approach first trains any generative model that yields a low-dimensional representation of an image. This may be done entirely off-the-shelf, enabling easy use of state-of-the-art models. Any model will suffice so long as it has a low-dimensional latent space that can accurately encode and manipulate an image.

Formally, let $\cE$ be an encoding procedure: the encoder of a VAE, for example, or a GAN inversion strategy like latent optimization. 
Recall the notation in Section~\ref{sec:background}, wherein a counterfactual on input $x$ to model $f$ aims to satisfy $f(x')=y'$. 
Encode $x$ with the latent vector $\cE(x) = z \in \mathbb{R}^d$. The generator $\cG$ maps $z$ back to image space, such that 
$$x^\text{Recon} = \cG(\cE(x))\approx x.$$

Given a model predicting $f(x)=y$, the counterfactual is an image $x'$ whose prediction is $y'$. Our approach modifies the latent representation $z$ to $z'$ in such a way that the reconstruction $x'=\cG(z)$ satisfies $f(x')=y'$. To do so, it runs gradient updates on the latent score itself until the score is attained. 
Suppose without loss of generality that $f$ outputs a scalar, and that $y'>f(x)$. 
With SGD and learning rate $\eta$, the update is
\begin{equation}\label{eq:grad}
    z \gets z +\eta\nabla_{z} f(\cG(z)). 
\end{equation}

\begin{algorithm}
\caption{Counterfactual Attacks}\label{alg:cf}
\begin{algorithmic}[1]
    \item[\textbf{Require:}] Input $x$, predictor $f$, target $y'>f(x)$, encoder $\cE$, generator $\cG$, learning rate $\eta$
    \item[\textbf{Ensure:}] Counterfactual $x'$ such that $f(x') \geq y'$
    \STATE $z \gets \cE(x)$
    \WHILE{$f(\cG(z)) \leq y'$}
        \STATE $z \gets z + \eta \nabla_{z} f(\cG(z))$ \hfill \COMMENT{Latent gradient ascent}
    \ENDWHILE
    \STATE $x' \gets \cG(z)$ \hfill \COMMENT{Generate counterfactual image}
    \STATE \textbf{return} $x'$
\end{algorithmic}
\end{algorithm}

Several adjustments may be made to the basic approach presented in Algorithm \ref{alg:cf}. 
Optimizers beyond SGD may be used, though the gradient computation is the same. 
If $y'<f(x)$, then the goal is to shrink the output of $f$, so gradient \textit{descent} is run rather than ascent. 
Further consider the case in which $f$ outputs a vector, as in multiclass classification. In this setting, modify Equation \eqref{eq:grad} to take only the coordinate of $f$ corresponding to the class of interest. $y'$ can be set to have a high value, i.e. reclassifying the input. 
Finally, it may be prudent to stop searching if no counterfactual is attained after a large number of iterations. This may be for computational convenience, or to exclude unrecognizably distant counterfactuals. 

The natural image manifold of GANs allows for smooth interpolation between latent variables \citep{zhu2016generative}. Our method is iterative, so intermediate steps may be visualized as well. This permits a smooth visual shift from original to counterfactual image.

\section{Feature Attributions for Image Counterfactuals}\label{sec:attributions}

We next introduce a framework to quantify image counterfactuals with feature attributions. These scores describe the important distinctions between the original image and its counterfactual. This approach applies for any counterfactual method, but has strong computational efficiency for our algorithm.

\subsection{General Schema}

In many image datasets, auxiliary feature labels are present. This is especially common for face datasets, tagging primarily categorical attributes like whether the person is smiling, has glasses, wears lipstick, etc.
We use these labels to quantify the content of counterfactual explanations with feature importance scores. 

Formally, let $\cA$ be the set of attributes for which labels are present, such as the aforementioned examples. 
Then, the images themselves $X$ are accompanied by attribute labels $Y_a\ \forall a \in \cA$.
Our approach first trains a machine learning model $g_a:\cX\rightarrow\cY_a$ to predict each attribute. These attribute predictors may input images in pixel space or a lower-dimensional representation.

We next consider the counterfactual explanation $x'$ for image $x$. This counterfactual may be obtained by \textit{any} method, not merely our Counterfactual Attack algorithm. Define each attribute's score as the change in prediction, normalized by the range if the feature is numeric. That is, the importance scores are 
\begin{equation}\label{eq:score}
    \phi_a(x,x') = \frac{g_a(x') - g_a(x)}{\text{diam}(\cY_a)}.
\end{equation}

The diameter of $\cY_a$ is the breadth of possible values it can take. For categorical data, this is just 1. For numeric features, finite limits may be known outright, for example a restricted set of pose angles. If not, it may be taken as the difference betwen the highest and lowest values of $g_a(x)$ observed in a background dataset.

As defined, the scores in Equation \eqref{eq:score} only apply to 1-dimensional outputs of $g_a(\cdot)$. This presents challenges for categorical attributes with more than two levels. To handle multi-class attribute models, one could analyze the levels in isolation, or aggregate them into a single attribute score, e.g. with the norm. An alternate approach would instead fit multiple binary classifiers.

The importance scores in Equation \eqref{eq:score} may be aggregated to produce global scores. 
A simple formula would take the sample average; or, one could normalize each image's scores by the sum across all attributes.
Either way, it is imperative to account for the fact that different counterfactuals move the model's predictions in different directions. In the binary setting, images predicted in the positive class are moved to the negative class, and vice versa. A straightforward approach would use absolute values:

\begin{equation}
    \psi_a = \sum_{i=1}^n \vert\phi_a(x_i, x_i')\vert.
\end{equation}

This approach loses a notion of directionality. One cannot discern from the global scores whether the presence of an attribute tends to move the prediction in the positive or negative direction. To accomplish this, one can weight the scores $\phi_a(x_i, x_i')$ by the direction of the counterfactual. 
\begin{equation}\label{eq:global}
    \psi_a = \sum_{i=1}^n \phi_a(x_i, x_i') s_i, \text{ where } s_i = 
    \begin{cases}
      1, & \text{if}\ f(x_i)<0.5 \\
      -1, & \text{otherwise}
    \end{cases}
\end{equation}

\subsection{Considerations for Counterfactual Attacks}

To ensure the importance scores are reliable, each attribute model $g_a(x)$ must generalize well.
This can only be achieved by neural networks for sophisticated image prediction tasks. Of course, training $\vert\cA\vert$ neural networks may come at a very high computational cost. Fortunately, when Counterfactual Attacks is used to generate counterfactuals, we can circumvent this computational burden almost entirely. 

\citet{interpreting_linear} showed that fitting linear classifiers in StyleGAN's latent space revealed meaningful feature directions. Having trained StyleGAN or a similar model for counterfactual explanations, it can be easily repurposed to fit these cheap attribute models. Moreover, far fewer samples may be necessary to fit them because the latent space is relatively low-dimensional. 

To do so, embed each image in the latent space with $z_i = \cE(x_i)$. Then, fit generalized linear models $h_a$ predicting each $y_a$ from $z$. The image-to-attribute scores are $g_a(x) = h_a(\cE(x))$. The standard choices of GLM are logistic regression for categorical data, linear regression for numeric data, and Poisson regression for count data.
GLM coefficients may also be regularized with Lasso and/or Ridge penalties. 

\section{Experiments}

We demonstrate the utility of Counterfactual Attacks on the MNIST and CelebA datasets. 
We did not benchmark it against a wide range of other methods, because its aim is not to produce the highest-quality counterfactuals; rather, it is to be the easiest to implement, a more subjective notion. Moreover, our methodology to depict expanations with feature importance scores is the first of its kind. 

However, Appendix \ref{apx:interfacegan} demonstrates the effectiveness of our feature attribution strategy on counterfactuals generated with InterFaceGAN \citep{interpreting_linear}. We also compared its counterfactuals to those from Counterfactual Attacks themselves. This is somewhat of an apples-to-oranges comparison, as InterFaceGAN uses a different classifier; nevertheless, we see that Counterfactual Attacks produces better counterfactual images on certain classes.

\subsection{MNIST}\label{sec:mnist}

To show its efficacy in simple settings, we ran Counterfactual Attacks on the MNIST dataset. These small grayscale images have dimensionality 28x28. We trained a straightforward neural network with 2 convolutional layers followed by 2 linear layers. Training stopped after only 1 epoch to ensure there were a reasonable number of misclassifications for failure analysis. The top-1 accuracy for both train and test sets was 97\%. 

For the manifold representation, we trained a variational autoencoder with a 64-dimensional latent space. Also a simple architecture, its encoder had 3 convolutional layers and 1 linear layer; the decoder reversed this process. To ensure its reconstructions were sparse, we added an L1 penalty to the loss. We also penalized the output of the Sobel operator in order to encourage smooth edges.

\begin{figure}[htbp]
  \centering
  \begin{subfigure}[b]{0.7\textwidth}
    \centering
    \includegraphics[width=\textwidth]{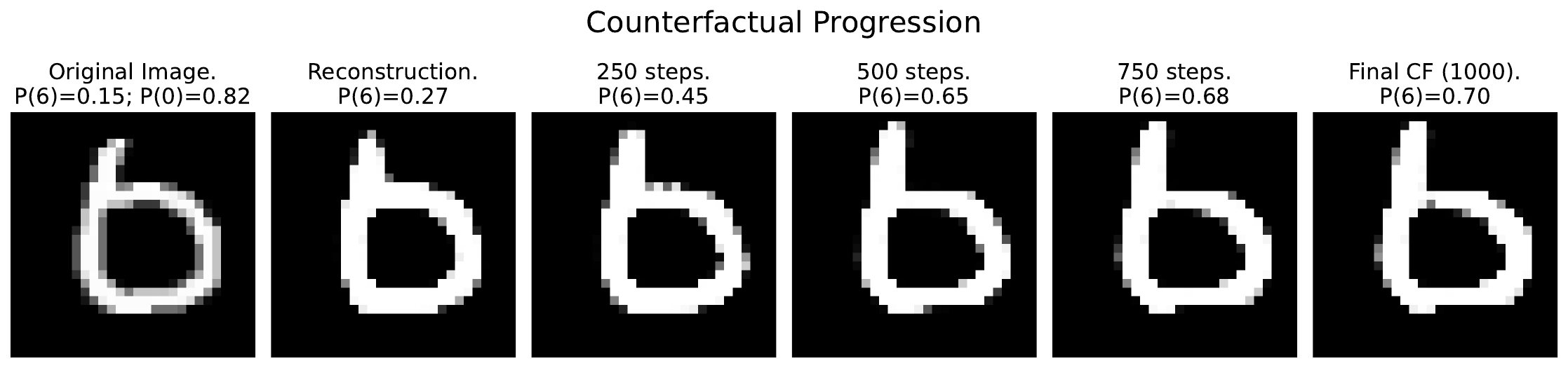}
    \caption{Counterfactual path towards correct classification.}
    \label{fig:mnist-prog}
  \end{subfigure}
  
  \vspace{1em}
  
  \begin{subfigure}[b]{0.7\textwidth}
    \centering
    \includegraphics[width=\textwidth]{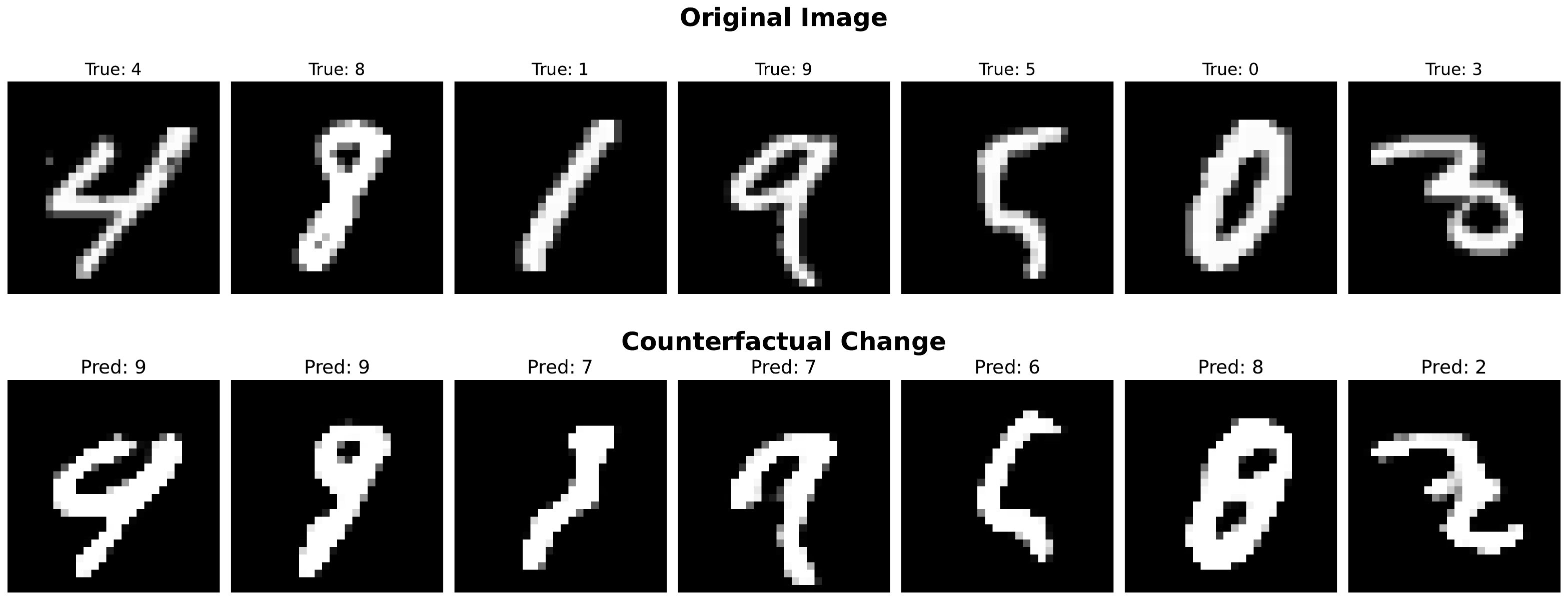}
    \caption{Transforming various digits.}
    \label{fig:mnist-trans}
  \end{subfigure}
  
  \caption{Running Counterfactual Attacks on MNIST dataset.}
  \label{fig:overall}
\end{figure}

Figure \ref{fig:overall} illustrates the capabilities of Counterfactual Attacks. Subfigure \ref{fig:mnist-prog} examines a failure mode in which the original digit 6 was misclassified by the network as a 0. We visualize the intermediate steps, as the model’s prediction is gradually nudged towards the correct class. These images capture how the digit’s structure is altered, with each step increasing the probability of predicting 6 by elongating the protrusion on the top left. This progression highlights how the method not only corrects misclassifications but also provides a transparent sequence that sheds light on the salient transformations. Appendix \ref{apx:mnist} displays a wide array of similar failure analyses. 

Subfigure \ref{fig:mnist-trans} showcases a different use case: transforming correctly-classified digits to a new number. For instance, a 4 is systematically turned into a 9 by curling and extending its left prong; a slender 8 is turned into a 9 by closing its bottom gap; and a 0 is turned into an 8 by squeezing its center and adding a connective bar. 
These results demonstrate our method’s flexibility in generating realistic digit morphs along the data manifold, while remaining coherent at each iteration.


\subsection{CelebA}

\begin{figure*}[!t]
    \centering
    \includegraphics[width=0.9\linewidth]{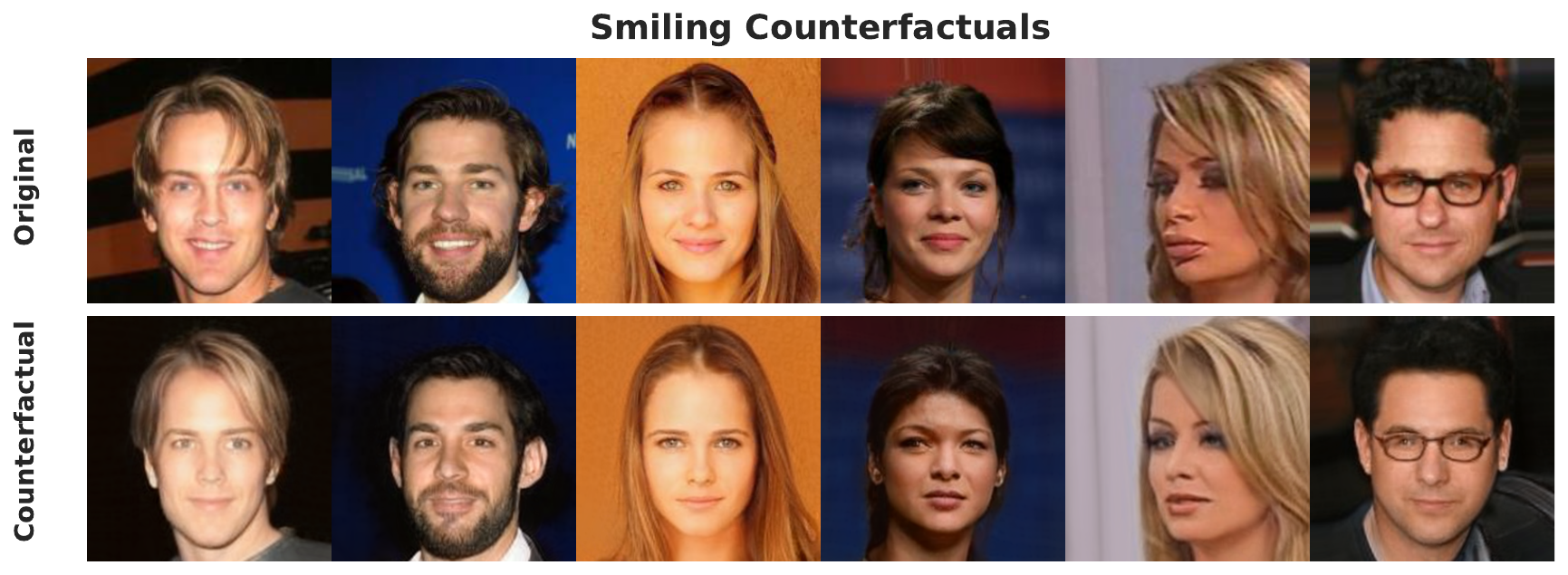}
    \caption{Counterfactuals for smiling classifier.}
    \label{fig:smiling}
\end{figure*}

We ran our methods on CelebA, a dataset of over 200,000 celebrity face images \citep{celeba}. The purpose of doing so was two-fold: To evaluate Counterfactual Attacks on a more complex dataset, as well as to demonstrate the capability of the feature attribution framework. 

CelebA contains 40 binary attribute labels, describing the physical features, accessories, and expressions of each image. We trained separate 5-layer CNNs on four of the attributes: Young, Attractive, Male, and Smiling. The first three are more subjective than descriptive attributes like baldness. As a result, they pose useful targets for our importance-scoring methodology, to elucidate what facial features drive predictions. 


For the generative model, we trained StyleGAN3 on CelebA \citep{stylegan3}. 
Works like InterFaceGAN \citep{interpreting_linear} and StylEx \citep{yossi_stylex} demonstrated the capacity of the StyleGAN models for semantic face editing. StyleGAN3 improved upon its predecessors by removing the capacity for aliasing, which had tied details to absolute image coordinates. 

After training, we represented each image in StyleSpace via the projection method of \citet{abdal2019image2stylegan}. 
Using \texttt{sklearn}, we fit logistic regression models with L2 penalties on 10,000 projected images. 
Counterfactuals were scored on the objective attributes, removing unhelpful or redundant features. Appendix \ref{apx:celeba_setup} provides a deeper account of our preprocessing and modeling choices.

Figure \ref{fig:smiling} shows various faces transformed by the smiling classifier. The top row shows the original image, followed by the altered counterfactual. In some cases, the counterfactual transformation is obvious as in the first three images. The latter three are much more subtle, yet nevertheless distinctive. Upon close inspection, it is clear that the 4th and 6th counterfactuals no longer smile, but the 5th does. Because smiling is a straightforward concept, we did not accompany the counterfactuals with importance scores. 

\begin{figure*}[!t]
    \centering
    \includegraphics[width=0.7\textwidth]{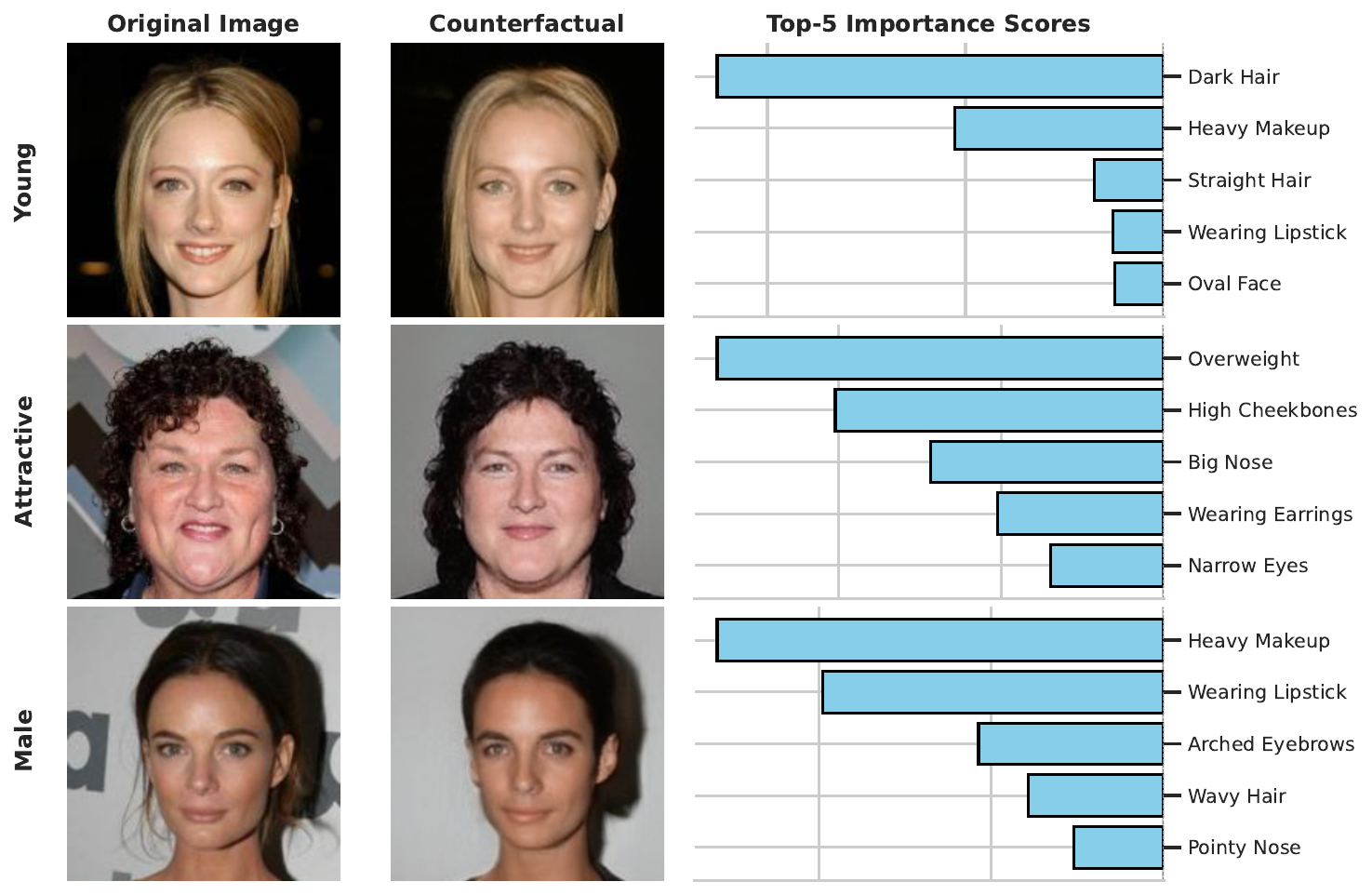}
    \caption{Counterfactual images with accompanying importance scores. Each row presents an individual counterfactual on a separate CNN classifier. All scores are negative, indicating the removal of their features.}
    \label{fig:labeled}
\end{figure*}

Figure \ref{fig:labeled} displays counterfactuals with top-5 importance scores on the attractiveness, age, and gender classifiers. In all cases, Counterfactual Attacks produces realistical counterfactuals. Furthermore, the counterfactuals are all accurately described by the importance scores. Examining the top-scoring features that accompany each counterfactual, the older woman has whiter hais and wears less makeup; the attractiveness classifier puts large emphasis on weight; and the transformation to male is in part defined by the lack of makeup, lipstick, and wavy hair. Appendix \ref{apx:celeba_pos_results} displays more examples of counterfactuals - both labeled as in Figure \ref{fig:labeled}, as well as for the Smiling classifier. 

\begin{figure*}
    \centering
    \includegraphics[width=\textwidth]{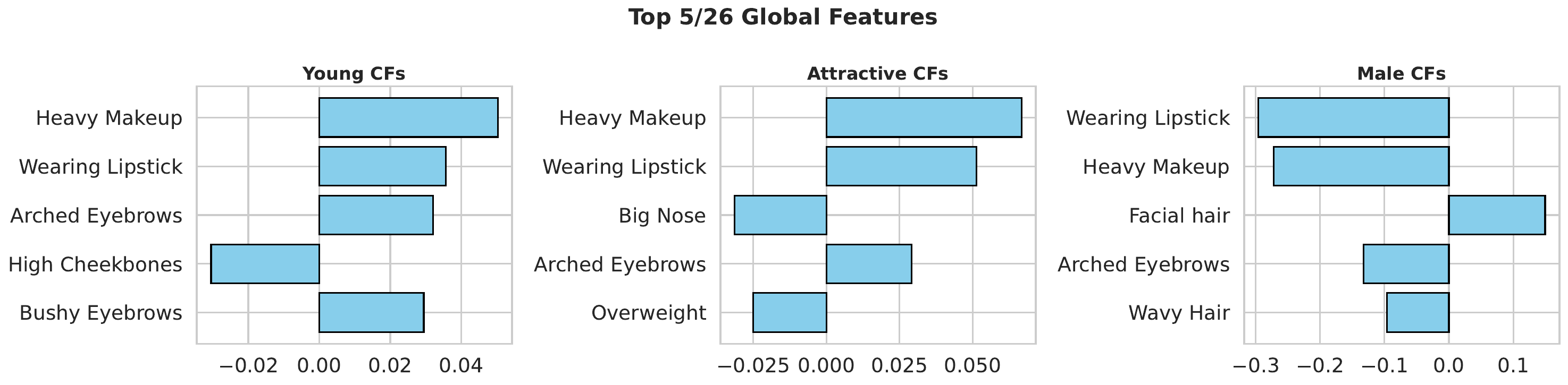}
    \caption{Global counterfactual explanations for three CelebA classifiers. The direction indicates whether the feature is added or removed.}
    \label{fig:global}
\end{figure*}

Figure \ref{fig:global} showcases global counterfactual explanations for these three classifiers. It averages 100 local counterfactuals according to Equation \eqref{eq:global}. The top-5 highlighted features are highly intuitive. The model characterizes youth, attractiveness, and femininity in celebrity photos with makeup, lipstick, and arched eyebrows. The model also associates young celebrities with bushy eyebrows and lower cheekbones. Indeed, eyebrows thin with age, and structural features like cheekbones may become more pronounced. Attractiveness is further characterized by being less overweight and having a slender nose, and masculinity with facial hair.

While powerful, our counterfactual techniques have their limitations, discussed in depth in Appendix \ref{apx:celeba_neg_results}. Firstly, the counterfactuals are only as good as the generative model they utilize. In our experiments with StyleGAN3, not every image projected neatly onto the latent space. This motivates the use of more sophisticated architectures, such as those with built-in encoders. Secondly, the labels must describe all possible transformations in order to be as useful as possible. While generally strong for youth and attractiveness, the provided CelebA attributes are limited for the gender classifier, as they do not include hair length. Wavy hair is the nearest proxy, hence its presence in Figures \ref{fig:labeled} and \ref{fig:global}.
Figures \ref{fig:poor_recons} and \ref{fig:hair_fail} visualize these two failure modes.

\section{Discussion}

Both counterfactuals and adversarial attacks make small changes to an image such that the prediction changes \citep{adversarial_attacks, cf_adv}.
We created a method that leverages this similarity, inspired by existing work in adversarial ML. 
Prior works on counterfactuals, and many adversarial methods, minimize a regularized loss function \citep{aversarial_og}. 
Ours is the first counterfactual method, however, to mimic the approach of crafting examples via iterative gradient ascent steps \citep{kurakin2017adversarialexamplesphysicalworld}.
The primary difference is that our ``attacks'' are performed within a low-dimensional latent space. This ensures that gradient steps walk along the natural image manifold -- presuming that the latent space is a good representation thereof. 

Our algorithm is more straightforward to implement than other methods for image counterfactuals. State-of-the-art generative models may be trained off-the-shelf, with no adjustment. Our experiments on CelebA used StyleGAN3, but other dimensionality reduction models may be used. Future work could explore the performance of Counterfactual Attacks across image autoencoders, GANs, LDMs, and DAEs. Moreover, because of the simplicity of gradient ascent, it is not necessary to tune regularization hyperparameters, as in prior work. 
Our approach may also satisfy the Rashomon Effect with a diverse set of counterfactuals, owing to the random noise inserted in the generation process. 

Another novel contribution of our work is its unification of image counterfactuals with feature importance scores. These scores may serve as a useful objective complement or substitute to visual inspection. They can also be aggregated for global attributions with Equation \eqref{eq:global} \citep{global_aggregations}. The feature rankings from these global scores can be verified with the retrospective procedure from \citet{us}. In general, global image counterfactuals are an underexplored area of research. The only prior work we are aware of is \citet{sobieski2024global}, which finds directions for global transformations in DAEs.

Global attributions facilitate an efficient, objective analysis of a model's overall behavior by removing the need to manually inspect individual counterfactuals. This capability is particularly valuable for identifying and understanding patterns behind systematic errors, such as recurring types of misclassifications. By summarizing the underlying reasons for these failures, global attributions can generate actionable insights that guide model debugging and improvement. For instance, if the model consistently misclassifies images of people wearing glasses, this insight could prompt the collection of more training data featuring glasses. Similarly, if poor lighting conditions lead to degraded performance, data augmentation techniques could be employed to simulate such conditions during training, thereby improving the model’s robustness.

In general, counterfactual explanations provide more actionable insights than perturbation- or gradient-based methods like LIME or GradCAM \citep{LIME,gradcam}. Rather than merely associating a score to pixels or patches, methods like Counterfactual Attacks provide examples of how to change a model's prediction. Using the paper's examples, a person could learn how to write zeros that looks less like sixes, or to present themselves more youthfully. This paper's works go a step further by also providing feature importance scores. These highlight the importance of factors that visual inspection might otherwise overlook. 

The importance scores we introduce describe the contents of the counterfactual post-hoc. Alternatively, one could investigate using them to generate the counterfactuals themselves. A simple way to do this would be to perform edits using the InterFaceGAN strategy until the desired prediction is reached. This may require composing edits on more than one feature. A more sophisticated approach would modify Counterfactual Attacks by projecting each gradient step onto the subspace defined by the attribute vectors. That way, the resulting counterfactual could be entirely accounted for by the labeled features.

\newpage
\bibliographystyle{apalike}
\bibliography{refs}

\newpage
\appendix
\section{MNIST}\label{apx:mnist}

\begin{figure}
    \centering
    \includegraphics[width=0.9\linewidth]{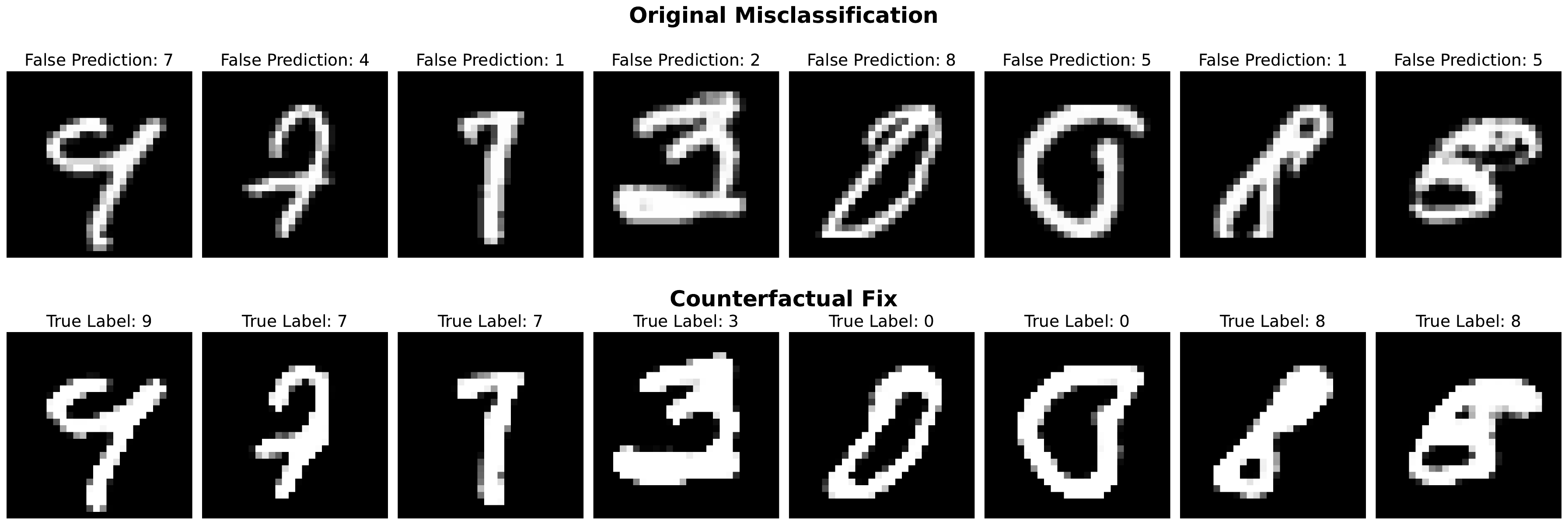}
    \caption{Counterfactuals that alter misclassified inputs to correct model predictions.}
    \label{fig:mnist-failures}
\end{figure}

Figure \ref{fig:mnist-failures} displays a number of misclassified inputs, and their accompnaying counterfactuals that steer them to the correct classification. We used the Adam optimizer \citep{kingma2015adam}, and a threshold of 0.5 for each image's counterfactual. The counterfactual approach was the same as in section \ref{sec:mnist}. The counterfactuals all make intuitive sense in explaining the model failures. For example, the 7 had been misclassified as a 1 because its tip dipped down; the 0 had been misclassified as an 8 because its top circled over; and the 8 had been misclassified as a 1 because its base was not connected. 

\section{CelebA}

\subsection{Experimental Setup}\label{apx:celeba_setup}

To train StyleGAN3, we used CelebA centered images of dimensionality 256x256. Training took 3 days on 4 A4000 GPUs, fine-tuning from weights pre-trained on the FlickrFaces (FFHQ) dataset. We trained the model to be translation and rotation equivariant, with a batch size of 32. 

The latent attribute predictors were fit using the default \texttt{sklearn} implementation of logistic regression. We combined several features into individual predictors:
\begin{itemize}
    \item \textbf{Facial hair} encompassed the results of ``sideburns,'' ``goattee,'' ``5 o'clock shadow,'' ``mustache,'' and ``no beard.'' Naturally, the ``no beard'' feature being present was marked as the absence of a beard.
    \item \textbf{Dark hair} encompassed ``black hair,'' ``brown hair,'' and the reverse of ''blond hair.''
    \item \textbf{Overweight} encompassed ``chubby'' and ``double chin.''
\end{itemize}

We also ignored attributes regarding expressions, image characteristics and more abstract features. This encompassed ``attractive,'' ``blurry,'' ``young,'' ``male,'' ``mouth slightly open,'' and ``smiling.'' Doing so was necessary to hone in on concrete facial features. 

Next, we trained neural networks for attractiveness, smiling, youth, and gender. The networks were all 5-layer CNNs, the first 3 of which were convolutional. All activation functions were ReLU, and dropout was applied before the final linear layer. We trained for 5 epochs on the full 200k-image dataset with an 80/20 train/test split. 
The resulting test accuracies were 80\% for attractive, 91\% for smiling, 90\% for gender, and 87\% for youth. In contrast, CelebA's class imbalances were roughly 50\% for attractive and smiling, 60\% women, and 75\% young. 

To generate counterfactuals, we ran Counterfactual Attacks with the Adam optimizer, initializing the learning rate to 0.01. We targeted a prediction of 0.75 if the original image was in predicted the negative class, i.e. if $f(x)<0.5$. Otherwise, positive predictions were transformed to a value of 0.25.

\subsection{Further Results}\label{apx:celeba_pos_results}

\begin{figure}
    \centering
    \includegraphics[width=0.9\linewidth]{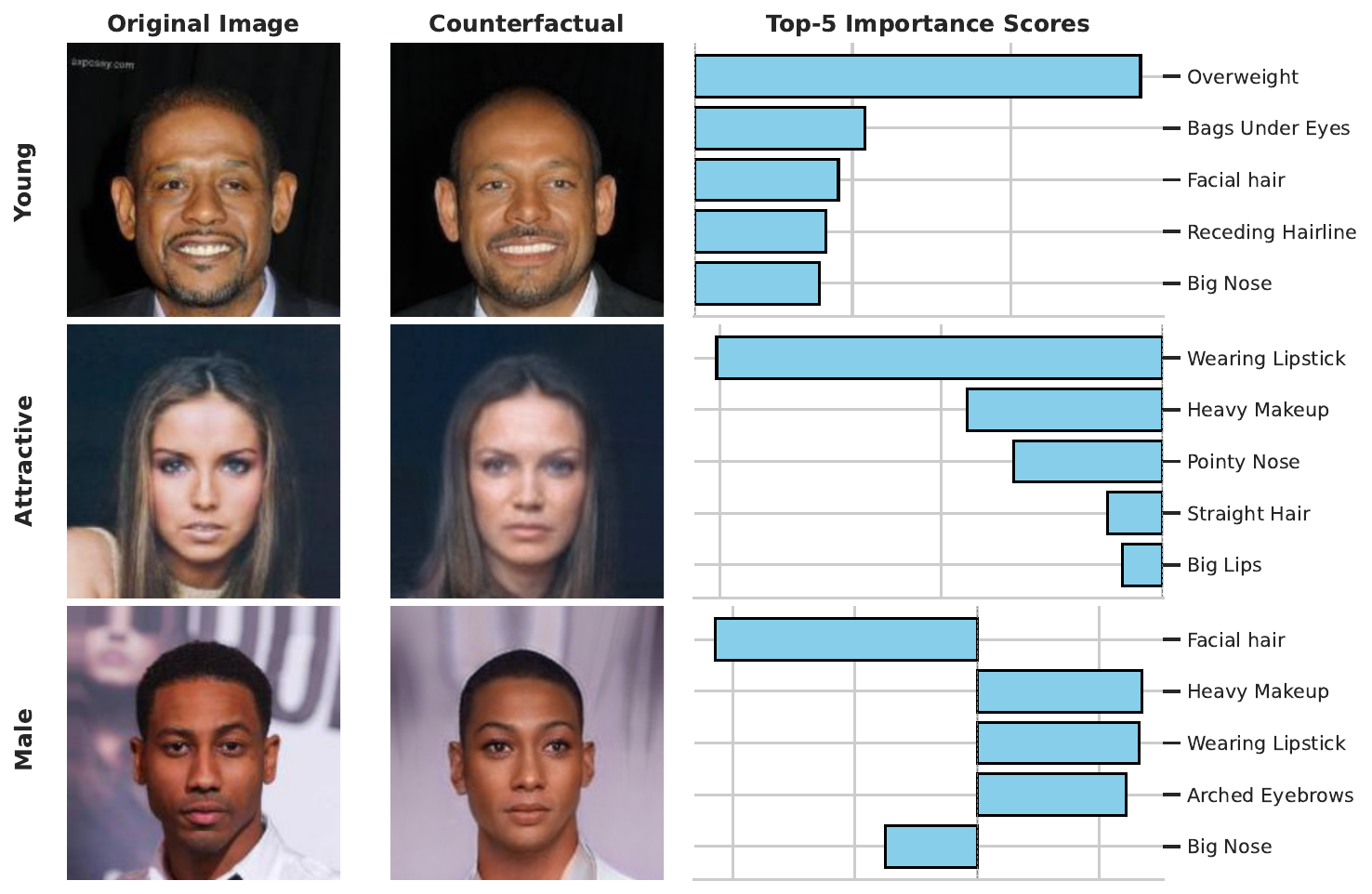}
    \caption{More examples of labeled counterfactuals.}
    \label{fig:labeled_cfs2}
\end{figure}

Figure \ref{fig:labeled_cfs2} displays three more counterfactuals with importance scores. From top to bottom, the older man's face is more rounded, and he clearly has a receding hairline. The woman scored as unattractive no longer wears lipstick or makeup, and her nose is less slim. The transformed woman has lost stubble, wears lipstick and makeup, and styles arched eyebrows.

\subsection{Limitations}\label{apx:celeba_neg_results}

\begin{figure}
    \centering
    \includegraphics[width=0.6\linewidth]{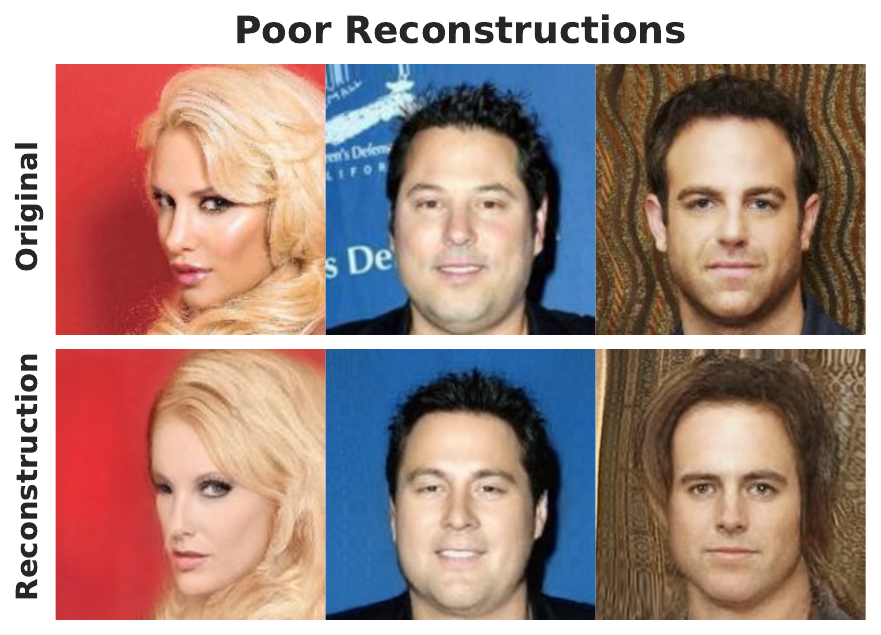}
    \caption{Poor reconstructions of various CelebA images.}
    \label{fig:poor_recons}
\end{figure}

Here, we showcase two potential failure modes of our method. Figure \ref{fig:poor_recons} displays three images whose StyleSpace projections do not represent the original image particularly well. Respectively, the reconstructions alter the profile, eyes, and hair of the three faces.
This could be remedied with a vast number of methods that better represent images in a low-dimensional space. Nevertheless, we present this to highlight that Counterfactual Attacks - or any counterfactual method, for that matter - will not perform well on images poorly represented in latent space. 

\begin{figure}
    \centering
    \includegraphics[width=0.8\linewidth]{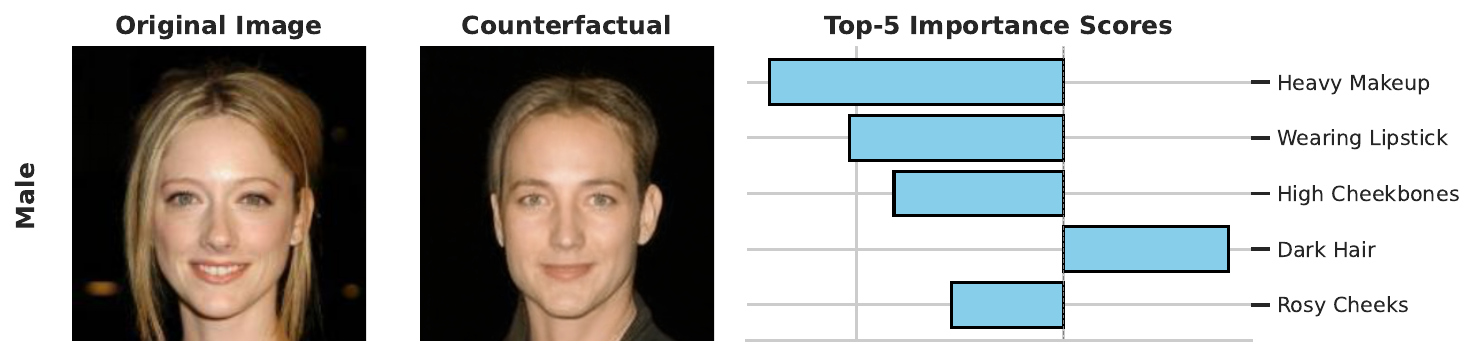}
    \caption{Gender counterfactual does not label the shortened hair length.}
    \label{fig:hair_fail}
\end{figure}

More specific to our contributions, Figure \ref{fig:hair_fail} shows a counterfactual on the perceived gender classifier. The realistic counterfactual retains most facial characteristics, but notably truncates the person's hair below the ears. However, this is not reflected in the shown feature importance scores, because hair length is not included in the attribute set. This demonstrates that the importance scores are only useful if they cover all reasonable explanations for the model's prediction to flip.

\section{Experimentation with InterFaceGAN}\label{apx:interfacegan}

We showcased the effectiveness of our feature attribution method to describe counterfactuals in general, not limited to Counterfactual Attacks. In particular, we applied it to counterfactual images generated with InterFaceGAN \citep{interpreting_linear}. 
Here, we describe InterFaceGAN in more detail than in Section \ref{sec:background}, and show how it can be repurposed to produce counterfactuals.

InterFaceGAN operates in the StyleSpace of a StyleGAN model. This space represents an image $x$ with an encoding $z$. To edit a face image with regards to some feature $y$, this method fits a linear classifier $f$ in StyleSpace.
Assume without loss of generality that the label of interest $y$ is binary; further assume it is predicted with logistic regression, outputting $\P(y=1\vert z)$. In this context, $\hat\beta$ and $\hat c$ are fit such that for logistic function $\sigma$,
\begin{equation}\label{eq:interfacegan}
    f(z) = \sigma(\langle z, \hat\beta\rangle+\hat c).
\end{equation}
Presuming the model in Equation \eqref{eq:interfacegan} is sufficiently accurate, $\hat\beta$ represents the \textit{direction} of the feature in StyleSpace. InterFaceGAN modifies the presence of the feature in the original image by moving its latent representation in that direction. For some value of $\eta\in\R$, InterFaceGAN generates the image 
$$\cG(z+\eta\hat\beta).$$
When $\eta$ is positive, the predicted probability that the feature is present increases. This aims to turn the feature \textit{on} if it is not present. Conversely, the feature may be removed from an image by moving in the negative direction of $\beta$. 

How much should the image be moved along this axis? While InterFaceGAN does not employ the language of counterfactuals, it can easily extended to this framework. Let $p$ be the predicted probability for image $z$ under the logistic regression model, and let $p'$ be the desired prediction of its counterfactual. We rearrange Equation \eqref{eq:interfacegan} to identify the value $\eta'$ that yields this probability:

\begin{align*}
    p'&=\sigma(\langle z+\eta'\beta, \beta\rangle +c)\\
    &=\sigma\!\left(\left(\langle z, \beta\rangle +c\right) + \eta'\lVert\beta\rVert_2^2\right)\\
    &= \sigma\!\left(\sigma^{-1}(p) + \eta'\lVert\beta\rVert_2^2\right)
\end{align*}
Defining the inverse logistic function $\sigma^{-1}(y)=\log(\frac{y}{1-y})$,
\begin{equation*}
    \sigma^{-1}(p') = \sigma^{-1}(p) + \eta\lVert\beta\rVert_2^2.
\end{equation*}
Solving for $\eta'$ reveals its value,
\begin{equation}\label{eq:eta}
    \eta'=\frac{\sigma^{-1}(p')-\sigma^{-1}(p)}{\lVert\beta\rVert_2^2}.
\end{equation}

On CelebA, we fit three latent logistic regression models for the Attractiveness, Youth, and Male attributes. Their Encoding a number of images, we identified values of $\eta'$ that would flip these models' predictions. As with the neural network classifiers, our counterfactuals changed positive predictions to 0.25, and negative predictions to 0.75. Each case produced a counterfactual latent vector, $z' = z+\eta'\hat\beta$. Their corresponding images were simply regenerated with $x'=\cG(z')$.

We employed our feature importance methodology to quantify the contents of these counterfactuals. To recap, our approach fits models $g_a$ for each attribute, and takes the change in prediction as the score. In this context, we used the same attribute predictors as for our Counterfactual Attacks experiments (Appendix \ref{apx:celeba_setup}). These were the logistic regression models that input images in latent space. (For our InterFaceGAN experiments, the latent space in question was the StyleSpace of StyleGAN3.)

\begin{figure}
    \centering
    \includegraphics[width=0.7\textwidth]{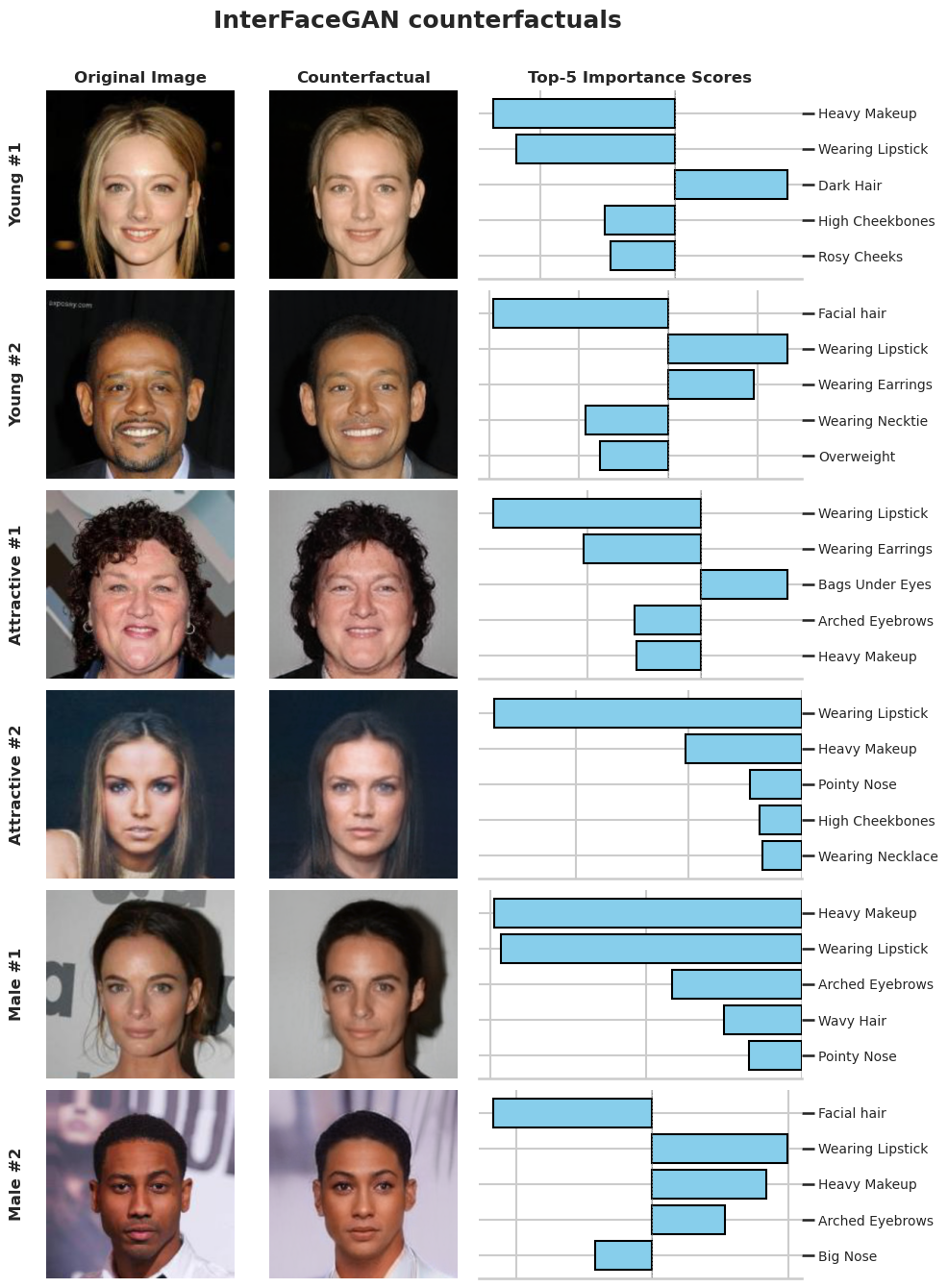}
    \caption{InterFaceGAN counterfactuals with accompanying importance scores from our methodology in Section \ref{sec:attributions}. Each row presents an individual counterfactual on a separate logistic regression classifier, fit in StyleSpace.}
    \label{fig:interfacegan_images}
\end{figure}

Figure \ref{fig:interfacegan_images} displays labeled InterFaceGAN counterfactuals with the same images as in Figures \ref{fig:labeled} and \ref{fig:labeled_cfs2}. Upon visual inspection, many of these counterfactuals seem to successfully flip the attribute label in question. Furthermore, the importance scores unanimously provide accurate description of their contents. For example, facial hair is clearly removed in the second and sixth counterfactuals. On all of the others, the absence or presence of lipstick is apparent. This indicates that our proposed method is capable of describing counterfactuals in general, and is not restricted to Counterfactual Attacks.

It is worth noting that Counterfactual Attacks algorithm seems to generate better counterfactuals for Youth and Attractiveness than InterFaceGAN. This may be explained by the descrepancies in the classifiers they interpret. For these features, the neural networks have roughly 3\% higher test accuracy. In fact, the attractiveness classifier gives roughly the same prediction for the original and counterfactual images in the third row.

The poorer performance of logistic regression could be due to Youth and Attractiveness being nonlinear features in StyleSpace. In fact, the non-linear direction learned by the Youth classifier seems to conflate older age with masculinity. Strikingly, the woman's hair is truncated; more subtly, the man loses facial hair and adds a hint of lipstick. Operating in image space, the neural networks have more flexible modeling capacity, and thus can avoid these linearity issues. 

\begin{figure}
    \centering
    \includegraphics[width=\textwidth]{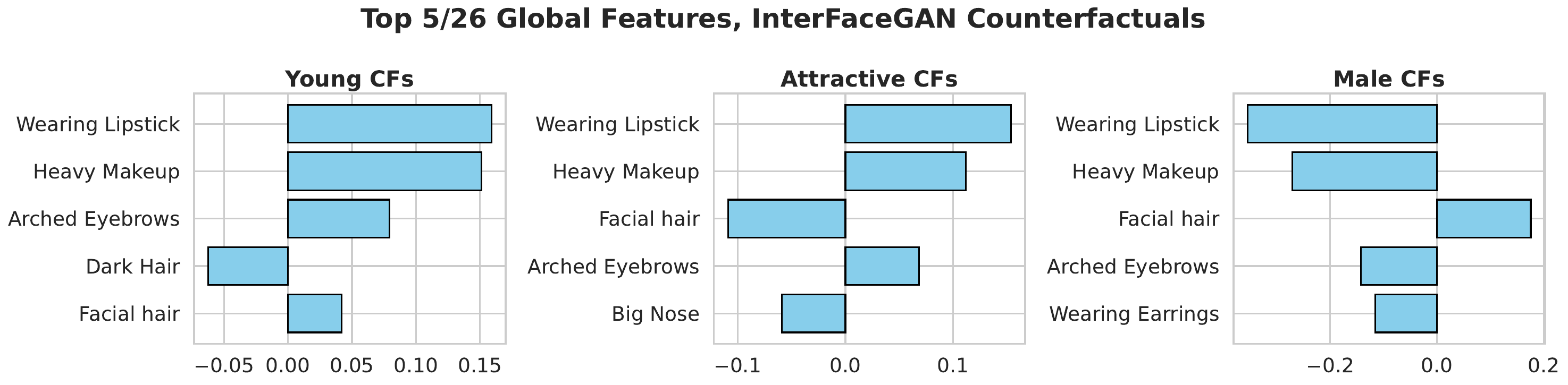}
    \caption{Top-5 global importance scores for InterFaceGAN counterfactual explanations. Three classifiers are logistic regression models fit in StyleSpace. The direction of each score indicates whether the feature is added or removed.}
    \label{fig:global_interfacegan}
\end{figure}

Finally, we aggregate local scores to produce global importance scores for InterFaceGAN counterfactuals. This follows an identical approach as for Counterfactual Attacks, in Figure \ref{fig:global}. On the same 100 images, we compute each latent-space counterfactual with InterFaceGAN, then take an average with Equation \eqref{eq:global}. Figure \ref{fig:global_interfacegan} visualizes the results. Intuitively, the model identifies lipstick, makeup, and arched eyebrows as characteristically feminine, and facial hair as masculine. 

These features are all important for the youth classifier as well, per its unfortunate conflation of age and gender. They also make the top-5 for the attractiveness classifier. While lipstick, makeup, and arched eyebrows are all sensible, the inclusion of facial hair suggests that classifier has learned to conflate attractiveness and femininity. This could explain in part why the classifier does not perform as well as the neural network. It also associates smaller noses with attractiveness.

\end{document}